\pdfoutput=1

\documentclass[11pt]{article}

\usepackage[table, dvipsnames]{xcolor}
\usepackage[]{ACL2023}
\usepackage{times}
\usepackage{latexsym}

\usepackage[T1]{fontenc}

\usepackage[utf8]{inputenc}

\usepackage{microtype}

\usepackage{inconsolata}

\usepackage{booktabs}
\usepackage{multirow}
\usepackage{graphicx}
\usepackage{amsmath}
\usepackage{xspace}

\newcommand{\lhigh}{{$L_\text{high}$}}
\newcommand{\llow}{{$L_\text{low}$}}

\title{OPEx: A Component-Wise Analysis of LLM-Centric Agents\\ in Embodied Instruction Following}

\author{
Haochen Shi$^1$, Zhiyuan Sun$^1$, Xingdi Yuan$^2$, Marc-Alexandre Côté$^2$\thanks{\ \ Equal advising.}, Bang Liu$^1$\footnotemark[1] \thanks{\ \ Canada CIFAR AI Chair.} \\
$^1$ Université de Montréal \& Mila, Montréal, Canada \\
$^2$ Microsoft Research, Montréal, Canada \\ 
\{haochen.shi, zhiyuan.sun, bang.liu\}@umontreal.ca, \\
\{eric.yuan, macote\}@microsoft.com
}

\begin{document}
\maketitle
\begin{abstract}

Embodied Instruction Following (EIF) is a crucial task in embodied learning, requiring agents to interact with their environment through egocentric observations to fulfill natural language instructions. Recent advancements have seen a surge in employing large language models (LLMs) within a framework-centric approach to enhance performance in embodied learning tasks, including EIF. Despite these efforts, there exists a lack of a unified understanding regarding the impact of various components—ranging from visual perception to action execution—on task performance. To address this gap, we introduce OPEx, a comprehensive framework that delineates the core components essential for solving embodied learning tasks: Observer, Planner, and Executor. Through extensive evaluations, we provide a deep analysis of how each component influences EIF task performance. Furthermore, we innovate within this space by deploying a multi-agent dialogue strategy on a TextWorld counterpart, further enhancing task performance. Our findings reveal that LLM-centric design markedly improves EIF outcomes, identify visual perception and low-level action execution as critical bottlenecks, and demonstrate that augmenting LLMs with a multi-agent framework further elevates performance.

\end{abstract}

\section{Introduction}


Embodied learning, particularly through tasks like Embodied Instruction Following (EIF)~\cite{shridhar2020alfred}, stands at the forefront of artificial intelligence research. EIF, where agents must interpret natural language instructions to navigate and act within their environment using egocentric observations, epitomizes the challenge of integrating cognitive understanding with physical action. This intersection is crucial for developing autonomous agents capable of nuanced interaction with complex, real-world environments, marking a significant stride towards more advanced and versatile AI systems. As the research community harnesses advancements in deep learning, we edge closer to this ambition~\cite{baker2022video, min2021film, inoue2022prompter, huang2022language}.


Traditional approaches to Embodied Instruction Following (EIF) often rely on expert-generated annotations, a process that can be both expensive and challenging to scale for real-world applications. In contrast, Large Language Models (LLMs), such as those cited in recent studies~\cite{inoue2022prompter, openai2023gpt, wei2022emergent, driess2023palm, touvron2023llama, huang2022language, huang2022inner, liang2022code, wang2023voyager, shinn2023reflexion, song2023llm}, have emerged as a potent alternative, showcasing exceptional capabilities in natural language understanding and generation. These models, enriched by extensive textual datasets, demonstrate significant common-sense reasoning abilities. As a result, there's a growing trend towards leveraging LLM-centric architectures for embodied learning tasks including EIF, which promise to simplify planning and execution tasks through a few-shot learning paradigm. However, despite their potential, the implementations of 
EIF systems introduce a variety of designs and components across different studies~\cite{min2021film,inoue2022prompter,song2023llm,blukis2022persistent,wang2023voyager,zhu2023ghost}. There remains a notable gap in systematically understanding how these disparate elements influence overall task performance, underscoring the need for a thorough analysis of LLM-centric methods within the context of EIF.



In addressing the complexities of Embodied Instruction Following (EIF), we introduce OPEx, a novel framework designed to systematically outline the essential components for mastering embodied learning tasks. OPEx is segmented into three core parts: Observer, Planner, and Executor. The Observer component is tasked with processing and interpreting sensory inputs, primarily visual, to construct an actionable understanding of the agent's immediate environment. The Planner dynamically devises strategic plans as subtasks to complete the tasks based on perceptual inputs, effectively bridging the gap between perception and action. Lastly, the Executor is responsible for implementing these plans with a skill library, which translates several re-useable skills into precise, context-aware actions within the environment, ensuring the agent's interactions are both relevant and goal-oriented. This tripartite structure provides a clear delineation of roles within the system, facilitating a granular analysis of how each contributes to the overarching performance of EIF tasks.

To understand the impact of each OPEx component on performance in EIF tasks, we conducted an in-depth analysis. By experimenting with different versions of the Observer, Planner, and Executor components, we assessed how each contributes to and influences overall success. This approach allowed us to identify the key attributes and design choices that enhance the system's ability to tackle complex embodied tasks, providing clear insights into optimizing embodied learning agents.

To further unlock the potential of LLMs in embodied learning, we eliminate the influence of visual perception and low-level action execution of the system utilizing a pure-text counterpart environment~\cite{shridhar2020alfworld} and further adopt a multi-agent dialogue strategy, splitting the instruction-following challenge into distinct reasoning and grounding roles handled by a reasoner agent and an actor agent, respectively. This dialogue-driven approach simplifies the task into decision-making processes, where both agents utilize world knowledge obtained from an explorer. This explorer gathers insights either through direct interaction with the environment or from human contributions, thereby enriching the collaborative problem-solving capabilities of the reasoner and actor with more grounded and informed decision-making.

Our experimental evaluation was conducted using the ALFRED~\cite{shridhar2020alfred} and ALFWorld~\cite{shridhar2020alfworld} benchmarks, providing a comprehensive testing ground for our extensive evaluation. The core analysis of our experiments underscores significant advancements: the LLM-centric approach notably enhances performance in EIF tasks. We pinpoint visual perception and low-level action execution as pivotal bottlenecks. Moreover, our results affirm that incorporating a multi-agent dialogue strategy into an LLM-centric task solver significantly boosts overall task performance on AFLWorld, showcasing the effectiveness of our proposed methodology in addressing the complexities of embodied learning tasks.

\section{Task Formulation}
\label{sec:formulation}

We benchmark our method with ALFRED~\cite{shridhar2020alfred} and its TextWorld counterpart ALFWorld~\cite{shridhar2020alfworld}. Both contain a set of environments associated with long-horizon household tasks specified by natural language instructions. 
The language instruction $L=\{L_\text{high}, L_\text{low}\}$ consists of instructions at two different levels: a high-level instruction goal \lhigh that summarizes the task and a sequence of low-level instructions \llow that depict the specific actions required.
At the time step $t$, ALFRED also provides a visual egocentric observation $V_t$ (text observation $\mathcal{L}_t$ if on ALFWorld) represents the world state $\mathcal{W}_t$.

Given the language instruction $L$ and an initial observation $V_0$ ($\mathcal{L}_0$ if on ALFWorld), the agent's objective is to generate an execution trajectory $\mathcal{T}=\left\langle V_0, a_0, V_1, a_1, \ldots, V_T, a_T\right\rangle$, where $a_t$ is the action taken by the agent at time step $t$, and $V_{t+1}$ is the observation of the world state $\mathcal{W}_{t+1}$ caused by that action. The action space $A$ can be categorized into two classes: navigation actions $A_N$ and interaction actions $A_I$, respectively\footnote{$A_N \in \{$ \textit{RotateRight}, \textit{RotateLeft}, \textit{MoveAhead}, \textit{LookUp}, \textit{LookDown}$\}$, $A_I \in \{$ \textit{PickupObject}, \textit{PutObject}, \textit{OpenObject}, \textit{CloseObject}, \textit{ToggleObjectOn}, \textit{ToggleObjectOff}, \textit{SliceObject}$\}$.}. In practice, we follow the setting of FILM~\cite{min2021film}, where the navigation actions $A_N$ are constrained to discrete values, and a pixel-wise interaction mask of the target object must be specified for interaction actions $A_I$.  There are seven types of household tasks, of which each episode is terminated either if an agent meets the goal conditions specified in $L$ (success) or reaches the maximum number of steps (fail).
See Appendix.~\ref{sec:alfred} for a task example in ALFRED.



\begin{figure}[t]
    \centering
    \includegraphics[width=0.5\textwidth]{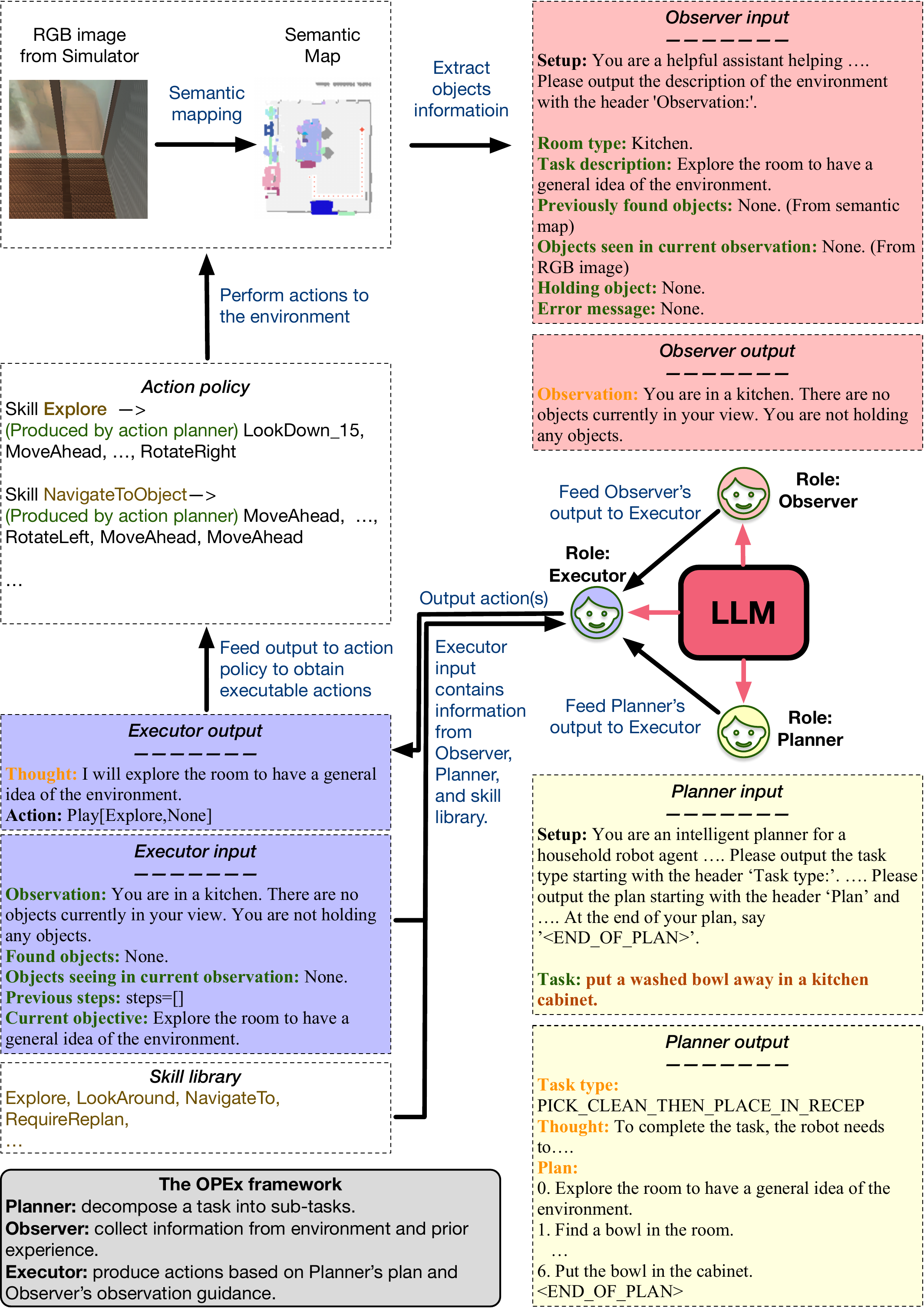}
    \caption{Overview of our OPEx framework. We will open-source the code after acceptance.}
    \label{fig:OPEx_framework}
    \vspace{-0.6cm}
\end{figure}

\section{Methodology}

We first provide an overview of the proposed LLM-centric framework (OPEx) in Figure \ref{fig:OPEx_framework}. 
OPEx consists of six components: (1) A \textit{semantic mapping module} to transform the egocentric visual observation into a semantic map; (2) An \textit{LLM-based planner} to decompose the specified language task instruction $L$ into subtasks $S=[S_0, S_1, ..., S_n]$; (3) An \textit{LLM-based observer} to gather information from the environment and depict the partially observed world state $\mathcal{W}_t$ at the current time step $t$ as natural language description $O^L_t$; (4) An \textit{LLM-based executor} to receive the world state description $O^L_t$ and select skill from a set of pre-defined skills to complete the current subtask $S_i$; (5) A \textit{skill library} $\mathcal{SL}=\{sl_0, sl_1, ...\}$ to store the skills manipulating the agent in the simulated environment (e.g, \textit{NavigateTo}, \textit{LookAround}, and \textit{Explore}); (6) A \textit{deterministic action policy} to convert the skills into low-level actions (e.g., \textit{RotateRight}). 

\paragraph{Semantic Mapping Module}
The goal of the semantic mapping module is to create a 2D semantic map $M_t$ from a top-down perspective (i.e., a map of explored areas, obstacles, and detected objects). At each time step $t$, this module receives the egocentric visual observation $V_t$ of the world state $\mathcal{W}_t$ as input, which is then processed into a depth map and instance segmentation using a UNet~\cite{ronneberger2015u} and 
 a MaskRCNN~\cite{he2017mask} (or ZoeDepth~\cite{bhat2023zoedepth} and  SOLQ~\cite{dong2021solq} as stronger perception models). Following FILM~\cite{min2021film}, we use the implementation from~\cite{blukis2022persistent} for UNet-based depth estimation and ~\cite{shridhar2020alfworld} for MaskRCNN-based instance segmentation. Then, a point cloud is constructed from the depth prediction and instance segmentation. Finally, the point cloud is binned into a voxel representation and then transformed into the 2D semantic map $M_t$ by summing over the height of the voxel representation, which is updated and aggregated over time steps. 
Due to the inherent difficulty in achieving a flawless perceptual model, the resulting semantic map $M_t$ often includes noise. This noise has the potential to exacerbate the challenges associated with locating navigational targets and subsequently affect the performance. To address such kind of issues, we introduce a supplementary semantic map denoted as $M_t'$, which aggregates the information from $M_t$ over successive time steps. The intuition resembles a form of majority voting: when an object is recognized as a fridge across more viewpoints than as a wall, its likelihood of being a fridge over a wall should be proportionally increased. The two semantic maps work in a cascading manner: when the agent tries to identify an object from the maps, the initial search is conducted within $M_t'$. $M_t$ is only utilized if the object cannot be located within $M_t'$.

\paragraph{LLM-based Planner}
The goal of the LLM-based planner is to decompose a specified language instruction $L$ into a sequence of subtasks $S=[S_0, S_1, ..., S_n]$. In practice, we utilize Chain-of-Though (CoT)~\cite{wei2022chain} to prompt GPT-4~\cite{openai2023gpt} with in-context learning. The corresponding prompt examples are demonstrated in the Appendix.

\paragraph{Example Selector} We have collected a set of prompt examples from 10 episodes within the training split for each of the 7 task types, amounting to a total of 70 episodes. As shown in~\cite{Liu2022kate}, choosing which in-context examples to add to the prompt can impact the overall performance. Therefore, we further apply an example selector to provide the LLM-based planner with the most relevant examples by ranking the examples based on the similarity of the input test case and the examples. In practice, we employ the example selector from LangChain~\cite{Chase_LangChain_2022}, which first ranks the examples based on the corresponding embeddings\footnote{We adopt the \textit{text-embedding-ada-002} embeddings provided by OpenAI.} that have the greatest cosine similarity with the inputs, then select top-$K$ examples for in-context learning. 

\paragraph{LLM-based Observer}


The goal of the LLM-based observer is to extract information from the environment feedback and the agent state, and present it in the form of a natural language description $O_t^L$ in a zero-shot manner. The rationale behind the design of the LLM-based observer is twofold: (1) to gather and render the state of the environment, enabling the tracking of environment dynamics across time steps and facilitating dynamic planning and acting; and (2) to summarize the information into a task-centric description, thereby safeguarding the LLM-based executor against distractions and hallucinations. The LLM-based observer is querying \textit{GPT-3.5-turbo} with the prompt format shown in the Appendix.

\paragraph{LLM-based Executor}
Given the current subtask $S_i$, the language description of the world state $O_t^L$ at time step $t$, the goal of the LLM-based executor is to complete the subtask $S_i$ by iteratively manipulating the agent in the environment with a set of pre-defined skills from a skill library $\mathcal{SL}$. In contrast to the LLM-based planner, which predominantly depends on the reasoning prowess of LLMs, the LLM-based executor is tasked with active engagement within the environment and acquiring an understanding of the environment dynamics (for instance, in ALFRED, objects can be cleaned by placing them into a sink basin and toggling on the faucet) from the feedback. To this end, inspired by ReAct~\cite{yao2022react}, we prompt the LLM-based executor (a \textit{GPT-4} model) to generate both reasoning traces and action plans (composed of skills in $\mathcal{SL}$), allowing for greater synergy between the two: reasoning traces help the model induce, track, and update action plans as well as handle exceptions, while actions allow it to interface with and gather additional information from the environment. 
The input for LLM-based Executor's prompt template is generally composed of the language-based observation $O_t^L$, found objects, objects detected in the current view, short-term memory of the action plan for the current subtask, which is cleared once the current subtask is finished, and the current subtask $S_i$. The LLM-based executor is required to generate both the reasoning traces (the ``Thought'' part in the Executor's output) and the action plans. The action space of the LLM-based executor is \{\emph{Play}, \emph{Finish}\}, where the action \emph{Play} is utilized to interact with the environment or request re-planning of the current plan $S$, and the action \emph{Finish} is used for finishing the action planning for the current subtask $S_i$. The action \emph{Play} receives two arguments as the inputs: [$\mathcal{SL}_i$, $\mathcal{ST}$] (e.g., Play[\textit{NavigateToObject},\textit{Table}]), where $\mathcal{SL}_i$ is the pre-defined skills in the skill library, and $\mathcal{ST}$ is the target argument of the corresponding skill action $\mathcal{SL}_i$.

\paragraph{Skill Library}
We design a skill library to empower the LLM-based executor with the following capabilities: (1) reasoning over language to track progress, handle exceptions or adjust the plan according to the situation; (2) acting to support the reasoning and collect information about the environment dynamics by controlling the agent. Apart from all the interaction actions $A_I$, we have designed several additional skills, including \textit{NavigateToObject}, \textit{Explore}, \textit{LookAround}, and \textit{RequireReplan}. The \textit{NavigateToObject} skill empowers the LLM-based executor with the capability to set the landmark-based navigation goal, it takes a found object in the room as the skill action target $\mathcal{ST}$. The \textit{Explore} skill enhances the LLM-based executor's ability to guide the agent in room exploration by sampling navigation goals from traversable areas, and it requires no skill action target. It is worth noting that we have an initial exploration heuristic for the first four calls of the \textit{Explore} skill, we set the four corners of the room with a higher exploration priority. The \textit{RequireReplan} provides the LLM-based executor with the capability to dynamically adjust the plan, improving the robustness to exceptions and producing more probability for it to learn from the environment dynamics. The \textit{LookAround} skill enables the LLM-based executor to manipulate the agent to look around the environment to get a more comprehensive observation of the room. 


\paragraph{Deterministic Action Policy}
Given the current instruction specified by the action plan [$\mathcal{SL}_i$ $\mathcal{ST}$] from the LLM-based executor, the deterministic action policy of OPEx outputs a navigation or interaction action based on a set of heuristics, which is quite similar to that of FILM. Both policies generally follow the following procedure: if the target object is observed in the semantic map, the closest instance is selected as the final navigation goal. Otherwise, the final navigation goal is set as the exploration navigation goal.
After goal determination, the agent employs the Fast Marching Method~\cite{sethian1996fast} to navigate towards the navigation goal. Additionally, when the target object is within the agent's egocentric visual range, the policy will try to conduct the interaction or adjust the position to prepare for the interaction action. The deterministic action policy of OPEx mainly differs from that of FILM in three aspects. Firstly, the deterministic action policy of OPEx is equipped with a slice replay heuristic, which tracks the location of successful execution of \textit{SliceObject} for easier going back. Secondly, instead of directly setting the location of the target object as the navigation goal, we sample a traversable location based on the distance to the target object as the navigation goal (noted as ``traversable goal heuristic''). Thirdly, instead of directly utilizing the semantic map $M_t$ to determine whether the target object is found and get the navigation goal for that object, we adopt the additional semantic map $M_t'$ to achieve this in the first place. If the target is not found in $M_t'$, the original semantic map $M_t$ is then utilized. We prioritize $M_t'$ as it is supposed to be more robust to the errors from the perception models.

\paragraph{Prior Knowledge Integration}
Due to the lack of prior knowledge of the specific environment, OPEx frequently fails even on ALFWorld where the impact of perception and action modules are ablated. For instance, OPEx may continuously fail for trying to pick up objects across various episodes due to the lack of the knowledge that agent can not directly hold more than 1 object in ALFRED. Furthermore, a system with a single agent trying to handle planning and grounding simultaneously often struggles to learn the optimal timing for switching between planning and grounding. To bridge the gap, we propose  improving OPEx by splitting the reasoning and grounding issues with a multi-agent dialogue strategy and marrying it with the world knowledge, which is obtained from an explorer by interacting with the environment or collecting human contributions. Specifically, we first deploy the agent to explore the ALFWorld environment and collect action-observation sequences $\{\mathcal{AO}_{i}\}$, where $\mathcal{AO}_i = [\mathcal{L}_0, a_0, \mathcal{L}_1, a_1, ..., \mathcal{L}_T]$. Then an LLM-based module or human is required to observe the action-observation sequences and summarize the world knowledge learned from $\{\mathcal{AO}_{i}\}$ as prior knowledge candidates $\{P_j\}$. After that, an LLM-based filter is applied on $\{P_j\}$ to eliminate contradictory and duplicated world knowledge, which results in the final set of world knowledge $\{P'_i\}$. Finally, the world knowledge $\{P'_i\}$ is integrated into the prompt templates of the multi-agent dialogue strategy, where a reasoner depicts general plans solving the task and the actor ground the plans as executable actions in the environment.


\section{Experiments and Discussion}
\subsection{Experiment Setup}
\paragraph{Evaluation Splits} 
The ALFRED benchmark consists of training, valid, and test sets. Both valid and test sets are composed of seen and unseen splits, where the unseen splits consist of rooms that do not appear in the training set. Following~\cite{yao2022react}, we evaluate our methods on 134 unseen evaluation games for the ALFWorld benchmark.

\paragraph{Evaluation Metrics}
Following~\cite{shridhar2020alfred, min2021film}, we report four evaluation metrics on AFLRED: (1) Success Rate (SR); 
(2) Goal Condition (GC), the ratio of goal conditions completed at the end of an episode; 
(3) path length weighted SR (PLWSR), the SR weighted by (path length of the expert trajectory)/(path length taken by the agent); 
(4) path length weighted GC (PLWGC), the GC weighted by the same factor. Following~\cite{yao2022react}, we report SR on ALFWorld.

\begin{table*}[]
\small
\centering
\begin{tabular}[width=1\textwidth]{llllllllll}
\toprule[1.5pt]
\multirow{2}{*}{Method} 
                        & \multicolumn{4}{c}{Test Seen} & & \multicolumn{4}{c}{Test Unseen} \\ \cline{2-5} \cline{7-10}
                        & PLWGC & GC    & PLWSR & SR   & & PLWGC  & GC     & PLWSR & SR    \\
                        \hline
\multicolumn{9}{l}{\textbf{ALFRED (High-level goal instructions only)}} \\
\hline
LAV~\cite{nottingham2021modular} & 13.18  &23.21 & 6.31  &13.35  & &10.47  &17.27  &3.12  &6.38\\
HLSM~\cite{blukis2022persistent}& 11.53 & 35.79 & 6.69 & 25.11  & & 8.45  &27.24  & 4.34  &16.29 \\
LGS-RPA~\cite{murray2022following} & \textbf{24.49}  &  41.71  & \textbf{16.65} &  33.01 & &  \textbf{20.01} & 38.55 & \textbf{12.92} & 27.80 \\
EPA~\cite{liu2022planning}&  3.47  &  44.14 & 2.56 &  39.96 & & 3.91  & 39.54 & 2.92 & 36.07 \\
LLM-Planner~\cite{song2023llm} & - & 24.57 &  - & 15.33 &  & - & 22.89 & - &  13.41 \\
FILM~\cite{min2021film}  & 14.17     & 36.15     & 10.39     & 25.77 & &  13.13  & 34.75 & 9.67  & 24.46\\
OPEx-S  &  20.13    &  \textbf{54.27}   &   13.64   & \textbf{43.51} & & 18.46  & \textbf{53.82} & 12.57 & \textbf{41.27} \\
\bottomrule[1.5pt]
\end{tabular}
\caption{Main Results on the test splits of ALFRED benchmark.}
\label{tab: OPEx_results}
\end{table*}

\paragraph{Compared Methods}
The compared methods on ALFRED include LAV~\cite{nottingham2021modular}, where the raw language and visual inputs are transformed into structured forms, with a separate ``action prediction module'' predicting the low-level actions; HLSM~\cite{blukis2022persistent}, a hierarchical approach that uses semantic voxel map state representation as a long-term
memory to solve long-horizon tasks; LGS-RPA~\cite{murray2022following},  which utilizes a Djikstra-based deterministic planner for navigation action generation and introduces landmark-guided search along with the reinforced pose adjustment for navigation goal searching and interaction action preparation respectively; EPA~\cite{liu2022planning}, a neural-symbolic approach with symbolic planning; LLM-Planner~\cite{song2023llm}, which simply prompts LLMs for task decomposition; FILM~\cite{min2021film}, which builds 2D semantic map and performs exploration with a semantic search policy. It is worth noting that there are also several works on the leaderboard reporting high performance that are not included in the comparison~\cite{inoue2022prompter, shridhar2020alfred, chen2023robogpt}, this is mainly because we focus on systematically outlining and evaluating the essential components for mastering EIF tasks, while we cannot find the description or available open-source resources of these works when we conduct the experiments. On the ALFWorld benchmark, apart from the variants of OPEx, we also introduce ReAct~\cite{yao2022react} for comparison to demonstrate the effectiveness of the proposed method.

\subsection{Experimental Results}
The main results are illustrated in Table~\ref{tab: OPEx_results}. When contrasting OPEx with the baseline FILM, it becomes evident that OPEx exhibits substantial improvement across two distinct environmental settings, encompassing both the goal condition (GC) and the success rate (SR). Notably, OPEx utilizes in-context learning on less than 10\% data used for FILMs' planner (Language Processor) training, while OPEx still significantly outperforms FILM. The observation that OPEx achieves 17.74\% and 16.78\% absolute gain in SR on test seen and unseen split respectively empirically demonstrates the effectiveness of the OPEx framework. However, it is also worth noting that the OPEx is inferior to FILM concerning the path length weighted metrics. This phenomenon could potentially be attributed to the deliberate choice of assigning a higher maximum number of failures to OPEx as compared to FILM. This choice typically leads to the average length of the resulting episodes. The rationale behind this decision was to encourage OPEx to undertake a more extensive exploration, thereby fostering the acquisition of skills in handling a broader range of exceptions arising from both uncommon scenarios and failures. On the other hand, the FILM utilizes two BERT models trained on the whole training set with the template assumption to conduct the task decomposition, while the LLM-based planner can achieve this goal with only a bunch of examples. This phenomenon shows that OPEx works with a much lower demand for in-domain data, making it more feasible in real-world scenarios, where the data collection could be more time-consuming and expensive. Furthermore, the FILM outputs low-level navigation and interaction actions solely with a deterministic policy, while OPEx introduces an LLM-based executor accompanying the deterministic policy to release LLMs' potential for robust language grounding and exception handling in the embodiment environment. Overall, the main results empirically demonstrate that it could be feasible to develop embodied experts with low demand for in-domain data by mining LLMs' potential for grounded planning and acting.

\subsection{Ablation Study and Analysis}
To further investigate the bottleneck of the system and the influence of different modules, we conduct several additional ablation studies.

\paragraph{Influence of perception models} We first conduct controlled experiments on the valid unseen split of the AFLRED dataset to study the influence of perception models. The corresponding results are illustrated in the first section of Table~\ref{tab: OPEx_ablation}, where OPEx-S denotes the OPEx with stronger perception models (fine-tuned ZoeDepth~\cite{bhat2023zoedepth} for depth prediction and SOLQ~\cite{dong2021solq} for instance segmentation), OPEx-P denotes the OPEx with perfect ground-truth depth prediction and instance segmentation. The performance gain from the improvement of perception models is very significant, indicating there is much room for improvement regarding the perception models in ALFRED.

\paragraph{Influence of action policies} As shown in the second section of Table~\ref{tab: OPEx_ablation}, we design and conduct another set of controlled experiments to study the influence of distinct deterministic action heuristics introduced. It can be seen from the table that setting the navigation goal inside the traversable area brings the most significant performance improvement, while slice replay brings marginal improvement. Besides, introducing the additional semantic map for robust landmark-based navigation goal searching brings moderate performance gain.

\paragraph{Influence of LLM-based modules} We first conduct controlled experiments on the validation unseen split of the dataset to study the influence of different modules. The corresponding results are illustrated in Table~\ref{tab: OPEx_ablation}. Significant performance degradation can be observed when the LLM-based planner is removed from the OPEx. This is probably attributed to the fact that the LLM-based executor is required to solely perform implicit long-term planning and grounded interaction simultaneously under this setting. The LLM-based observer is designed to gather information and help the LLM-based executor to focus on task-relevant information by summarizing collected information and filtering out the task-irrelevant counterparts. However, the ablation study shows that the performance gain brought by the LLM-based observer is marginal. This observation can be caused by several possible reasons, including (1) GPT-4's strong long text processing capability mitigates the needs of such kind of LLM-based observer; (2) the collected information from ALFRED is typically not too large/complex to cause severe distraction or hallucination of the LLM-based executor; (3) the observer utilizes zero-shot prompt, better prompts may need to be designed.

\paragraph{Influence of prior knowledge} To further investigate the role of decision-making modules in EIF agents, we conduct experiments on ALFWorld to eliminate the impact of perception models and action policies. The corresponding results are illustrated in the fourth section of Tabel~\ref{tab: OPEx_ablation}, where OPEx-L denotes the OPEx with prior knowledge learned from the environment and OPEx-H denotes the OPEx with prior knowledge provided by humans. With the observation that the system performance grows as the quality of the prior knowledge increases, this can be empirically explained by the intuition that decomposing EIF tasks via a collaborative multi-agent dialogue strategy helps intra-agent specialization and inter-agent cooperation. Besides, the intuition that the grounded prior knowledge prevents the agents from repetitive errors and facilitates grounded exception handling might also contribute to the results. Furthermore, the performance improvement of ReAct also empirically demonstrates the effectiveness of the proposed method.

\begin{table}[]
\small
\centering
\begin{tabular}[width=1\textwidth]{llllllllll}
\toprule[1.5pt]
\multirow{2}{*}{Method} & \multicolumn{4}{c}{Valid Uneen} \\ \cline{2-5}  & PLWGC & GC    & PLWSR & SR \\
\hline
\rowcolor{green!20} \multicolumn{5}{l}{\textbf{Influence of perception models}} \\
\rowcolor{green!20} OPEx & 13.48 & 48.61 & 9.08 & 35.91 \\
\rowcolor{green!20} OPEx-S & 16.52 & 51.28 & 11.38 &  40.80 \\
\rowcolor{green!20} OPEX-P & 23.72 & 66.17 & 17.43 & 59.43 \\
\hline
\rowcolor{yellow!20} \multicolumn{5}{l}{\textbf{Influence of action policies}} \\
\rowcolor{yellow!20}OPEx & 13.48 & 48.61 & 9.08 & 35.91 \\
\rowcolor{yellow!20}-semantic map $M'_t$ & 12.37 & 45.41 & 8.06 & 36.17 \\
\rowcolor{yellow!20}-slice replay & 12.64 & 45.25 & 8.35 & 37.39 \\
\rowcolor{yellow!20}-traversable goal & 11.77 & 43.49 & 7.09 & 34.50 \\
\hline
\rowcolor{orange!20}\multicolumn{5}{l}{\textbf{Influence of LLM-based modules}} \\
\rowcolor{orange!20}OPEx & 13.48 & 48.61 & 9.08 & 35.91 \\
\rowcolor{orange!20}-Planner & 8.10 & 40.16 & 5.72 & 30.57 \\
\rowcolor{orange!20}-Observer & 13.41 & 45.62 & 8.58 & 37.76 \\
\hline
\rowcolor{blue!10}\multicolumn{5}{l}{\textbf{Influence of prior knowledge (On ALFWorld)}} \\
\rowcolor{blue!10}ReAct & - & - & - & 66\\
\rowcolor{blue!10}OPEx &  - & - & -  & 73 \\
\rowcolor{blue!10}OPEx-L &  - & - & - & 78 \\
\rowcolor{blue!10}OPEx-H & - & - & - & 84\\
\bottomrule[1.5pt]
\end{tabular}
\caption{Ablation Studies of OPEx. OPEx-S denotes the OPEx with stronger perception models, OPEx-P denotes the OPEx with perfect ground-truth depth prediction and instance segmentation, OPEx-L denotes the OPEx with prior knowledge learned from the environment, and OPEx-H denotes the OPEx with prior knowledge provided by humans.}
\label{tab: OPEx_ablation}
\end{table}

\begin{table}[t]
\small
\centering
\begin{tabular}[width=0.5\textwidth]{lllll}
\toprule[1.5pt]
Method        & SR    & GC    & PLWSR & PLWGC \\ \hline
OPEx           & 38.12 & 46.13 & 9.03 & 13.45 \\
FILM           & 0.00 & 12.18&  0.00 & 2.78 
\\
\bottomrule[1.5pt]
\end{tabular}
\caption{Performance comparison with the baseline trained on same amount of data.}
\label{tab: OPEx data efficiency}
\vspace{-0.6cm}
\end{table}

\paragraph{Low demand for in-domain data}
To assess the efficiency of in-domain data usage, we conducted experiments comparing OPEx with the baseline FILM. The FILM is trained on identical data used for in-context learning of OPEx. The corresponding results are presented in Table~\ref{tab: OPEx data efficiency}. Our findings indicate that OPEx markedly outperforms FILM across all evaluation metrics in the unseen validation split. Empirically, this suggests that OPEx requires significantly less in-domain data compared to FILM. This controlled study underscores the potential of addressing embodied tasks through an LLM-based framework. This framework achieves low in-domain data demand EIF by integrating feedback mechanisms, closed-loop grounded planning, and action, harmonized with the reasoning and common sense capabilities of Large Language Models (LLMs). Moreover, it also prompts our further exploration into the trade-off between in-domain data efficiency and inference overhead, inspiring future directions, such as devising agents that adeptly integrate both common sense and in-domain knowledge in a data-efficient manner.

\paragraph{Error mode analysis} We conduct the error mode analysis of OPEx on the valid unseen split. The corresponding statics are shown in Table~\ref{tab:OPEx_errormode}. While our approach to calculate the statistics may vary from that of FILM, we have also incorporated FILM's statistics from the original paper~\cite{min2021film} for reference. Since we conduct the task decomposition with the LLM-based planner, which does not follow the template assumption, we don't have statistics on language processing errors. As shown in the table, the goal object not found error typically account for a great ratio of all kinds of error, indicating both FILM and OPEx suffer from imperfect perception models. Besides, the interactive exploration of the LLM-based executor and the deterministic heuristics probably brings a lower error rate of collisions and the error caused by the target object in a closed receptacle.

\begin{table}[t]
\small
\centering
\begin{tabular}[width=0.5\textwidth]{lll}
\toprule[1.5pt]
Error mode                  & FILM  & OPEx  \\ \hline
Goal object not found       & 26.07 & 27.36 \\   
Interaction failures        & 8.54  & 12.80 \\
Collisions                  & 11.00 & 9.84 \\ 
Object in closed receptacle & 16.16 & 11.61 \\
Language processing error   & 24.54 & - \\
Others                      & 13.69 &  38.39\\
\bottomrule[1.5pt]
\end{tabular}
\caption{Error mode analysis of OPEx on the valid unseen split.}
\label{tab:OPEx_errormode}
\vspace{-0.6cm}
\end{table}

\section{Related Work}
\paragraph{LLM-based Agents}
Significant progress has been made for LLM-based agents, which mainly focus on the following three aspects. \textit{LLM-centric Planning} utilizes LLMs to generate plans in dynamic environments. It can be further categorized into methods planning without feedback~\cite{huang2022language, fan2022minedojo, yao2022react, huang2022inner, xiang2023language, lin2023swiftsage} and approaches planning with feedback from environment, human, and model~\cite{wang2023voyager, zhu2023ghost, shinn2023reflexion, wang2023describe, rana2023sayplan, guan2023leveraging, kim2023language}. 
\textit{LLM-oriented Memory} stores information from the environment and boosts agents' capabilities of experience accumulation and self-evolving to facilitate future actions.~\cite{significant-gravitas2023auto-gpt, shinn2023reflexion, wang2023voyager, majumder2023clin, wang2023jarvis}
\textit{LLM-centric Action Policy} grounds the plans made by the agent into feasible action space~\cite{huang2022language, schick2023toolformer} Notably, our LLM-centric agent differs from Voyager~\cite{wang2023voyager} and GITM~\cite{zhu2023ghost} by mitigating the instruction grounding problem with dynamically adjusted plans from various granularity based on task-centric feedback from the environment. 

\paragraph{Instruction Following in Embodied Environment} Prior work on EIF in embodied environments can be categorized into two classes: \textit{Supervisely trained end-to-end or modular-based methods} that are eager for supervision signals from training data and hard to generalize due to the lack of abstraction and reasoning abilities~\cite{shridhar2020alfred, suglia2021embodied, pashevich2021episodic, blukis2022persistent, singh2020moca, liu2022lebp, min2021film, sharma2021skill}, and \textit{LLM-based methods} that utilizes LLMs' reasoning capability~\cite{inoue2022prompter, song2023llm}. Different from Prompter~\cite{inoue2022prompter} and LLM-Planner~\cite{song2023llm}, which introduce LLMs only for target location finding and dynamic task decomposition, our method is an LLM-centric framework and decouples reasoning tasks for decision masking problem with multiple LLM-based roles, where the LLMs build the plan, adjust the plan, and ground the plan into structured action spaces. Besides, our method evolves based on the feedback, providing promising future research directions, including human-in-the-loop learning, multi-source feedback mixing and refining, etc.

\section{Conclusion}

We introduce OPEx, an LLM-centric framework tailored for Embodied Instruction Following (EIF), and undertake extensive evaluations to dissect the influence of its distinct components. Building on this foundation, we further improve OPEx by integrating world knowledge with a multi-agent dialogue strategy to further harness LLMs' potential in addressing EIF challenges. Our comprehensive analysis reveals that an LLM-centric design significantly enhances EIF performance, pinpointing visual perception and low-level action execution as crucial bottlenecks. Additionally, our findings demonstrate that integrating a multi-agent dialogue mechanism within LLMs markedly boosts their effectiveness, offering promising directions for future research in embodied learning.


\section*{Limitations}
While our study introduces the OPEx framework and a dialogue-based mechanism for solving EIF tasks, it is not without its limitations. First, the reliance on large language models (LLMs) and the complexity of the multi-agent system introduce challenges in interpretability and computational efficiency. These models demand considerable resources by extensively communicating with ChatGPT, which might limit their applicability in resource-constrained environments. Second, our experiments are conducted within the confines of the ALFRED and ALFWORLD benchmarks, which, while comprehensive, may not encompass all possible real-world scenarios an embodied agent might encounter. Third, the integration of visual perception and action execution as identified bottlenecks suggests that further refinement in these areas is necessary to achieve truly seamless and adaptive embodied AI systems. Future work should aim to address these limitations, exploring more efficient model architectures, broader applicability across diverse environments, and enhanced methods for achieving naturalistic human-agent interaction.

\section*{Ethical Concerns}
We do not foresee an immediate ethical or societal impact resulting from our work.
However, as an LLM application, we acknowledge that OPEx could in some way be affected by various types of hallucinations introduced by the LLMs.
We therefore urge researchers and practitioners to use our proposed framework in a mindful way, especially when deploying such LLM-centric agents in real world applications.. 

\bibliography{anthology,custom}

\begin{thebibliography}{47}
\expandafter\ifx\csname natexlab\endcsname\relax\def\natexlab#1{#1}\fi

\bibitem[{Baker et~al.(2022)Baker, Akkaya, Zhokov, Huizinga, Tang, Ecoffet, Houghton, Sampedro, and Clune}]{baker2022video}
Bowen Baker, Ilge Akkaya, Peter Zhokov, Joost Huizinga, Jie Tang, Adrien Ecoffet, Brandon Houghton, Raul Sampedro, and Jeff Clune. 2022.
\newblock Video pretraining (vpt): Learning to act by watching unlabeled online videos.
\newblock \emph{Advances in Neural Information Processing Systems}, 35:24639--24654.

\bibitem[{Bhat et~al.(2023)Bhat, Birkl, Wofk, Wonka, and M{\"u}ller}]{bhat2023zoedepth}
Shariq~Farooq Bhat, Reiner Birkl, Diana Wofk, Peter Wonka, and Matthias M{\"u}ller. 2023.
\newblock Zoedepth: Zero-shot transfer by combining relative and metric depth.
\newblock \emph{arXiv preprint arXiv:2302.12288}.

\bibitem[{Blukis et~al.(2022)Blukis, Paxton, Fox, Garg, and Artzi}]{blukis2022persistent}
Valts Blukis, Chris Paxton, Dieter Fox, Animesh Garg, and Yoav Artzi. 2022.
\newblock A persistent spatial semantic representation for high-level natural language instruction execution.
\newblock In \emph{Conference on Robot Learning}, pages 706--717. PMLR.

\bibitem[{Chase(2022)}]{Chase_LangChain_2022}
Harrison Chase. 2022.
\newblock \href {https://github.com/hwchase17/langchain} {{LangChain}}.

\bibitem[{Chen et~al.(2023)Chen, Cui, Chen, Tan, Zhang, Zhao, and Wang}]{chen2023robogpt}
Yaran Chen, Wenbo Cui, Yuanwen Chen, Mining Tan, Xinyao Zhang, Dongbin Zhao, and He~Wang. 2023.
\newblock Robogpt: an intelligent agent of making embodied long-term decisions for daily instruction tasks.
\newblock \emph{arXiv preprint arXiv:2311.15649}.

\bibitem[{Dong et~al.(2021)Dong, Zeng, Wang, Zhang, and Wei}]{dong2021solq}
Bin Dong, Fangao Zeng, Tiancai Wang, Xiangyu Zhang, and Yichen Wei. 2021.
\newblock Solq: Segmenting objects by learning queries.
\newblock \emph{Advances in Neural Information Processing Systems}, 34:21898--21909.

\bibitem[{Driess et~al.(2023)Driess, Xia, Sajjadi, Lynch, Chowdhery, Ichter, Wahid, Tompson, Vuong, Yu et~al.}]{driess2023palm}
Danny Driess, Fei Xia, Mehdi~SM Sajjadi, Corey Lynch, Aakanksha Chowdhery, Brian Ichter, Ayzaan Wahid, Jonathan Tompson, Quan Vuong, Tianhe Yu, et~al. 2023.
\newblock Palm-e: An embodied multimodal language model.
\newblock \emph{arXiv preprint arXiv:2303.03378}.

\bibitem[{Fan et~al.(2022)Fan, Wang, Jiang, Mandlekar, Yang, Zhu, Tang, Huang, Zhu, and Anandkumar}]{fan2022minedojo}
Linxi Fan, Guanzhi Wang, Yunfan Jiang, Ajay Mandlekar, Yuncong Yang, Haoyi Zhu, Andrew Tang, De-An Huang, Yuke Zhu, and Anima Anandkumar. 2022.
\newblock Minedojo: Building open-ended embodied agents with internet-scale knowledge.
\newblock \emph{arXiv preprint arXiv:2206.08853}.

\bibitem[{Guan et~al.(2023)Guan, Valmeekam, Sreedharan, and Kambhampati}]{guan2023leveraging}
Lin Guan, Karthik Valmeekam, Sarath Sreedharan, and Subbarao Kambhampati. 2023.
\newblock Leveraging pre-trained large language models to construct and utilize world models for model-based task planning.
\newblock \emph{arXiv preprint arXiv:2305.14909}.

\bibitem[{He et~al.(2017)He, Gkioxari, Doll{\'a}r, and Girshick}]{he2017mask}
Kaiming He, Georgia Gkioxari, Piotr Doll{\'a}r, and Ross Girshick. 2017.
\newblock Mask r-cnn.
\newblock In \emph{Proceedings of the IEEE international conference on computer vision}, pages 2961--2969.

\bibitem[{Huang et~al.(2022{\natexlab{a}})Huang, Abbeel, Pathak, and Mordatch}]{huang2022language}
Wenlong Huang, Pieter Abbeel, Deepak Pathak, and Igor Mordatch. 2022{\natexlab{a}}.
\newblock Language models as zero-shot planners: Extracting actionable knowledge for embodied agents.
\newblock In \emph{International Conference on Machine Learning}, pages 9118--9147. PMLR.

\bibitem[{Huang et~al.(2022{\natexlab{b}})Huang, Xia, Xiao, Chan, Liang, Florence, Zeng, Tompson, Mordatch, Chebotar et~al.}]{huang2022inner}
Wenlong Huang, Fei Xia, Ted Xiao, Harris Chan, Jacky Liang, Pete Florence, Andy Zeng, Jonathan Tompson, Igor Mordatch, Yevgen Chebotar, et~al. 2022{\natexlab{b}}.
\newblock Inner monologue: Embodied reasoning through planning with language models.
\newblock \emph{arXiv preprint arXiv:2207.05608}.

\bibitem[{Inoue and Ohashi(2022)}]{inoue2022prompter}
Yuki Inoue and Hiroki Ohashi. 2022.
\newblock Prompter: Utilizing large language model prompting for a data efficient embodied instruction following.
\newblock \emph{arXiv preprint arXiv:2211.03267}.

\bibitem[{Kim et~al.(2023)Kim, Baldi, and McAleer}]{kim2023language}
Geunwoo Kim, Pierre Baldi, and Stephen McAleer. 2023.
\newblock Language models can solve computer tasks.
\newblock \emph{arXiv preprint arXiv:2303.17491}.

\bibitem[{Liang et~al.(2022)Liang, Huang, Xia, Xu, Hausman, Ichter, Florence, and Zeng}]{liang2022code}
Jacky Liang, Wenlong Huang, Fei Xia, Peng Xu, Karol Hausman, Brian Ichter, Pete Florence, and Andy Zeng. 2022.
\newblock Code as policies: Language model programs for embodied control.
\newblock \emph{arXiv preprint arXiv:2209.07753}.

\bibitem[{Lin et~al.(2023)Lin, Fu, Yang, Brahman, Huang, Bhagavatula, Ammanabrolu, Choi, and Ren}]{lin2023swiftsage}
Bill~Yuchen Lin, Yicheng Fu, Karina Yang, Faeze Brahman, Shiyu Huang, Chandra Bhagavatula, Prithviraj Ammanabrolu, Yejin Choi, and Xiang Ren. 2023.
\newblock Swiftsage: A generative agent with fast and slow thinking for complex interactive tasks.
\newblock \emph{arXiv preprint arXiv:2305.17390}.

\bibitem[{Liu et~al.(2022{\natexlab{a}})Liu, Liu, He, and Yang}]{liu2022lebp}
Haoyu Liu, Yang Liu, Hongkai He, and Hangfang Yang. 2022{\natexlab{a}}.
\newblock Lebp--language expectation \& binding policy: A two-stream framework for embodied vision-and-language interaction task learning agents.
\newblock \emph{arXiv preprint arXiv:2203.04637}.

\bibitem[{Liu et~al.(2022{\natexlab{b}})Liu, Shen, Zhang, Dolan, Carin, and Chen}]{Liu2022kate}
Jiachang Liu, Dinghan Shen, Yizhe Zhang, Bill Dolan, Lawrence Carin, and Weizhu Chen. 2022{\natexlab{b}}.
\newblock \href {https://doi.org/10.18653/v1/2022.deelio-1.10} {What makes good in-context examples for gpt-3?}
\newblock \emph{Proceedings of Deep Learning Inside Out (DeeLIO 2022): The 3rd Workshop on Knowledge Extraction and Integration for Deep Learning Architectures}.

\bibitem[{Liu et~al.(2022{\natexlab{c}})Liu, Palacios, and Muise}]{liu2022planning}
Xiaotian Liu, Hector Palacios, and Christian Muise. 2022{\natexlab{c}}.
\newblock A planning based neural-symbolic approach for embodied instruction following.
\newblock \emph{Interactions}, 9(8):17.

\bibitem[{Majumder et~al.(2023)Majumder, Mishra, Jansen, Tafjord, Tandon, Zhang, Callison-Burch, and Clark}]{majumder2023clin}
Bodhisattwa~Prasad Majumder, Bhavana~Dalvi Mishra, Peter Jansen, Oyvind Tafjord, Niket Tandon, Li~Zhang, Chris Callison-Burch, and Peter Clark. 2023.
\newblock Clin: A continually learning language agent for rapid task adaptation and generalization.
\newblock \emph{arXiv preprint arXiv:2310.10134}.

\bibitem[{Min et~al.(2021)Min, Chaplot, Ravikumar, Bisk, and Salakhutdinov}]{min2021film}
So~Yeon Min, Devendra~Singh Chaplot, Pradeep Ravikumar, Yonatan Bisk, and Ruslan Salakhutdinov. 2021.
\newblock Film: Following instructions in language with modular methods.
\newblock \emph{arXiv preprint arXiv:2110.07342}.

\bibitem[{Murray and Cakmak(2022)}]{murray2022following}
Michael Murray and Maya Cakmak. 2022.
\newblock Following natural language instructions for household tasks with landmark guided search and reinforced pose adjustment.
\newblock \emph{IEEE Robotics and Automation Letters}, 7(3):6870--6877.

\bibitem[{Nguyen et~al.(2021)Nguyen, Suganuma, and Okatani}]{nguyen2021look}
Van-Quang Nguyen, Masanori Suganuma, and Takayuki Okatani. 2021.
\newblock Look wide and interpret twice: Improving performance on interactive instruction-following tasks.
\newblock \emph{arXiv preprint arXiv:2106.00596}.

\bibitem[{Nottingham et~al.(2021)Nottingham, Liang, Shin, Fowlkes, Fox, and Singh}]{nottingham2021modular}
Kolby Nottingham, Litian Liang, Daeyun Shin, Charless~C Fowlkes, Roy Fox, and Sameer Singh. 2021.
\newblock Modular framework for visuomotor language grounding.
\newblock \emph{arXiv preprint arXiv:2109.02161}.

\bibitem[{OpenAI(2023)}]{openai2023gpt}
R~OpenAI. 2023.
\newblock Gpt-4 technical report.
\newblock \emph{arXiv}, pages 2303--08774.

\bibitem[{Pashevich et~al.(2021)Pashevich, Schmid, and Sun}]{pashevich2021episodic}
Alexander Pashevich, Cordelia Schmid, and Chen Sun. 2021.
\newblock Episodic transformer for vision-and-language navigation.
\newblock In \emph{Proceedings of the IEEE/CVF International Conference on Computer Vision}, pages 15942--15952.

\bibitem[{Rana et~al.(2023)Rana, Haviland, Garg, Abou-Chakra, Reid, and Suenderhauf}]{rana2023sayplan}
Krishan Rana, Jesse Haviland, Sourav Garg, Jad Abou-Chakra, Ian Reid, and Niko Suenderhauf. 2023.
\newblock Sayplan: Grounding large language models using 3d scene graphs for scalable task planning.
\newblock \emph{arXiv preprint arXiv:2307.06135}.

\bibitem[{Ronneberger et~al.(2015)Ronneberger, Fischer, and Brox}]{ronneberger2015u}
Olaf Ronneberger, Philipp Fischer, and Thomas Brox. 2015.
\newblock U-net: Convolutional networks for biomedical image segmentation.
\newblock In \emph{Medical Image Computing and Computer-Assisted Intervention--MICCAI 2015: 18th International Conference, Munich, Germany, October 5-9, 2015, Proceedings, Part III 18}, pages 234--241. Springer.

\bibitem[{Schick et~al.(2023)Schick, Dwivedi-Yu, Dess{\`\i}, Raileanu, Lomeli, Zettlemoyer, Cancedda, and Scialom}]{schick2023toolformer}
Timo Schick, Jane Dwivedi-Yu, Roberto Dess{\`\i}, Roberta Raileanu, Maria Lomeli, Luke Zettlemoyer, Nicola Cancedda, and Thomas Scialom. 2023.
\newblock Toolformer: Language models can teach themselves to use tools.
\newblock \emph{arXiv preprint arXiv:2302.04761}.

\bibitem[{Sethian(1996)}]{sethian1996fast}
James~A Sethian. 1996.
\newblock A fast marching level set method for monotonically advancing fronts.
\newblock \emph{proceedings of the National Academy of Sciences}, 93(4):1591--1595.

\bibitem[{Sharma et~al.(2021)Sharma, Torralba, and Andreas}]{sharma2021skill}
Pratyusha Sharma, Antonio Torralba, and Jacob Andreas. 2021.
\newblock Skill induction and planning with latent language.
\newblock \emph{arXiv preprint arXiv:2110.01517}.

\bibitem[{Shinn et~al.(2023)Shinn, Cassano, Labash, Gopinath, Narasimhan, and Yao}]{shinn2023reflexion}
Noah Shinn, Federico Cassano, Beck Labash, Ashwin Gopinath, Karthik Narasimhan, and Shunyu Yao. 2023.
\newblock \href {http://arxiv.org/abs/2303.11366} {Reflexion: Language agents with verbal reinforcement learning}.

\bibitem[{Shridhar et~al.(2020{\natexlab{a}})Shridhar, Thomason, Gordon, Bisk, Han, Mottaghi, Zettlemoyer, and Fox}]{shridhar2020alfred}
Mohit Shridhar, Jesse Thomason, Daniel Gordon, Yonatan Bisk, Winson Han, Roozbeh Mottaghi, Luke Zettlemoyer, and Dieter Fox. 2020{\natexlab{a}}.
\newblock Alfred: A benchmark for interpreting grounded instructions for everyday tasks.
\newblock In \emph{Proceedings of the IEEE/CVF conference on computer vision and pattern recognition}, pages 10740--10749.

\bibitem[{Shridhar et~al.(2020{\natexlab{b}})Shridhar, Yuan, C{\^o}t{\'e}, Bisk, Trischler, and Hausknecht}]{shridhar2020alfworld}
Mohit Shridhar, Xingdi Yuan, Marc-Alexandre C{\^o}t{\'e}, Yonatan Bisk, Adam Trischler, and Matthew Hausknecht. 2020{\natexlab{b}}.
\newblock Alfworld: Aligning text and embodied environments for interactive learning.
\newblock \emph{arXiv preprint arXiv:2010.03768}.

\bibitem[{Significant-gravitas et~al.(2023)}]{significant-gravitas2023auto-gpt}
Significant-gravitas et~al. 2023.
\newblock Significant-gravitas/auto-gpt: An experimental open-source attempt to make gpt-4 fully autonomous.
\newblock https://github.com/Significant-Gravitas/Auto-GPT.
\newblock Open-Source Software.

\bibitem[{Singh et~al.(2020)Singh, Bhambri, Kim, Mottaghi, and Choi}]{singh2020moca}
Kunal~Pratap Singh, Suvaansh Bhambri, Byeonghwi Kim, Roozbeh Mottaghi, and Jonghyun Choi. 2020.
\newblock Factorizing perception and policy for interactive instruction following.
\newblock \emph{arXiv preprint arXiv:2012.03208}.

\bibitem[{Song et~al.(2023)Song, Wu, Washington, Sadler, Chao, and Su}]{song2023llm}
Chan~Hee Song, Jiaman Wu, Clayton Washington, Brian~M Sadler, Wei-Lun Chao, and Yu~Su. 2023.
\newblock Llm-planner: Few-shot grounded planning for embodied agents with large language models.
\newblock In \emph{Proceedings of the IEEE/CVF International Conference on Computer Vision}, pages 2998--3009.

\bibitem[{Suglia et~al.(2021)Suglia, Gao, Thomason, Thattai, and Sukhatme}]{suglia2021embodied}
Alessandro Suglia, Qiaozi Gao, Jesse Thomason, Govind Thattai, and Gaurav Sukhatme. 2021.
\newblock Embodied bert: A transformer model for embodied, language-guided visual task completion.
\newblock \emph{arXiv preprint arXiv:2108.04927}.

\bibitem[{Touvron et~al.(2023)Touvron, Lavril, Izacard, Martinet, Lachaux, Lacroix, Rozi{\`e}re, Goyal, Hambro, Azhar et~al.}]{touvron2023llama}
Hugo Touvron, Thibaut Lavril, Gautier Izacard, Xavier Martinet, Marie-Anne Lachaux, Timoth{\'e}e Lacroix, Baptiste Rozi{\`e}re, Naman Goyal, Eric Hambro, Faisal Azhar, et~al. 2023.
\newblock Llama: Open and efficient foundation language models.
\newblock \emph{arXiv preprint arXiv:2302.13971}.

\bibitem[{Wang et~al.(2023{\natexlab{a}})Wang, Xie, Jiang, Mandlekar, Xiao, Zhu, Fan, and Anandkumar}]{wang2023voyager}
Guanzhi Wang, Yuqi Xie, Yunfan Jiang, Ajay Mandlekar, Chaowei Xiao, Yuke Zhu, Linxi Fan, and Anima Anandkumar. 2023{\natexlab{a}}.
\newblock Voyager: An open-ended embodied agent with large language models.
\newblock \emph{arXiv preprint arXiv:2305.16291}.

\bibitem[{Wang et~al.(2023{\natexlab{b}})Wang, Cai, Liu, Jin, Hou, Zhang, Lin, He, Zheng, Yang et~al.}]{wang2023jarvis}
Zihao Wang, Shaofei Cai, Anji Liu, Yonggang Jin, Jinbing Hou, Bowei Zhang, Haowei Lin, Zhaofeng He, Zilong Zheng, Yaodong Yang, et~al. 2023{\natexlab{b}}.
\newblock Jarvis-1: Open-world multi-task agents with memory-augmented multimodal language models.
\newblock \emph{arXiv preprint arXiv:2311.05997}.

\bibitem[{Wang et~al.(2023{\natexlab{c}})Wang, Cai, Liu, Ma, and Liang}]{wang2023describe}
Zihao Wang, Shaofei Cai, Anji Liu, Xiaojian Ma, and Yitao Liang. 2023{\natexlab{c}}.
\newblock Describe, explain, plan and select: Interactive planning with large language models enables open-world multi-task agents.
\newblock \emph{arXiv preprint arXiv:2302.01560}.

\bibitem[{Wei et~al.(2022{\natexlab{a}})Wei, Tay, Bommasani, Raffel, Zoph, Borgeaud, Yogatama, Bosma, Zhou, Metzler et~al.}]{wei2022emergent}
Jason Wei, Yi~Tay, Rishi Bommasani, Colin Raffel, Barret Zoph, Sebastian Borgeaud, Dani Yogatama, Maarten Bosma, Denny Zhou, Donald Metzler, et~al. 2022{\natexlab{a}}.
\newblock Emergent abilities of large language models.
\newblock \emph{arXiv preprint arXiv:2206.07682}.

\bibitem[{Wei et~al.(2022{\natexlab{b}})Wei, Wang, Schuurmans, Bosma, Xia, Chi, Le, Zhou et~al.}]{wei2022chain}
Jason Wei, Xuezhi Wang, Dale Schuurmans, Maarten Bosma, Fei Xia, Ed~Chi, Quoc~V Le, Denny Zhou, et~al. 2022{\natexlab{b}}.
\newblock Chain-of-thought prompting elicits reasoning in large language models.
\newblock \emph{Advances in Neural Information Processing Systems}, 35:24824--24837.

\bibitem[{Xiang et~al.(2023)Xiang, Tao, Gu, Shu, Wang, Yang, and Hu}]{xiang2023language}
Jiannan Xiang, Tianhua Tao, Yi~Gu, Tianmin Shu, Zirui Wang, Zichao Yang, and Zhiting Hu. 2023.
\newblock Language models meet world models: Embodied experiences enhance language models.
\newblock \emph{arXiv preprint arXiv:2305.10626}.

\bibitem[{Yao et~al.(2022)Yao, Zhao, Yu, Du, Shafran, Narasimhan, and Cao}]{yao2022react}
Shunyu Yao, Jeffrey Zhao, Dian Yu, Nan Du, Izhak Shafran, Karthik Narasimhan, and Yuan Cao. 2022.
\newblock React: Synergizing reasoning and acting in language models.
\newblock \emph{arXiv preprint arXiv:2210.03629}.

\bibitem[{Zhu et~al.(2023)Zhu, Chen, Tian, Tao, Su, Yang, Huang, Li, Lu, Wang et~al.}]{zhu2023ghost}
Xizhou Zhu, Yuntao Chen, Hao Tian, Chenxin Tao, Weijie Su, Chenyu Yang, Gao Huang, Bin Li, Lewei Lu, Xiaogang Wang, et~al. 2023.
\newblock Ghost in the minecraft: Generally capable agents for open-world enviroments via large language models with text-based knowledge and memory.
\newblock \emph{arXiv preprint arXiv:2305.17144}.

\end{thebibliography}
\bibliographystyle{acl_natbib}

\clearpage
\newpage
\appendix

\section{Task Example in ALFRED}
\label{sec:alfred}
\begin{figure}[t]
    \centering
    \includegraphics[width=0.48\textwidth]{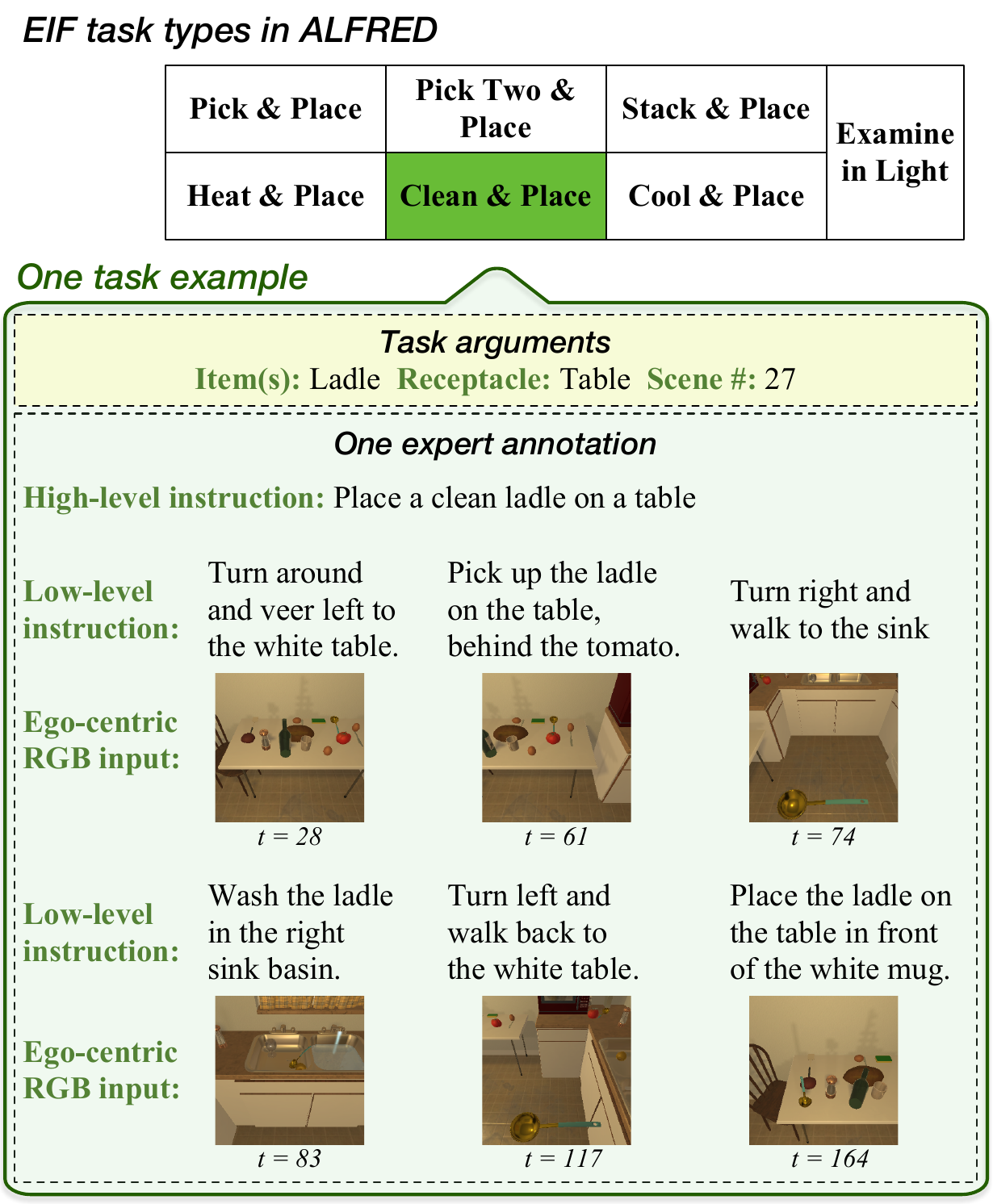}
    \caption{Example of a Clean \& Place task in ALFRED.}
    \label{fig:alfred}
    \vspace{-0.6cm}
\end{figure}

As shown in Figure~\ref{fig:alfred}, the ALFRED benchmark~\cite{shridhar2020alfred} contains a set of environments associated with long-horizon household tasks specified by natural language instructions. 
As shown in Figure~\ref{fig:alfred}, the language instruction $L=\{L_\text{high}, L_\text{low}\}$ consists of instructions at two different levels: a high-level instruction goal \lhigh that summarizes the task and a sequence of low-level instructions \llow that depict the specific actions required.
At the time step $t$, ALFRED also provides a visual egocentric observation $V_t$ represents the world state $\mathcal{W}_t$.
There are seven types of household tasks in ALFRED, namely Pick \& Place, Stack \& Place, Pick Two \& Place, Clean \& Place, Heat \& Place, Cool \& Place, and Examine in Light. An episode is terminated either if an agent meets the goal conditions specified in $L$ (success) or reaches the maximum number of steps (fail).

\section{Full Results on AFLRED}
The experiment on ALFRED under two different settings are illustrated in Table~\ref{tab: OPEx_full_results}.
\begin{table*}[]
\small
\centering
\begin{tabular}[width=1\textwidth]{llllllllll}
\toprule[1.5pt]
\multirow{2}{*}{Method} 
                        & \multicolumn{4}{c}{Test Seen} & & \multicolumn{4}{c}{Test Unseen} \\ \cline{2-5} \cline{7-10}
                        & PLWGC & GC    & PLWSR & SR   & & PLWGC  & GC     & PLWSR & SR    \\
                        \hline
\multicolumn{9}{l}{\textbf{High-level Goal Instruction + Low-level step-by-step instructions }} \\
\hline
Seq2Seq~\cite{shridhar2020alfred} &  6.27 & 9.42 & 2.02 & 3.98& &  4.26&  7.03 & 0.08&  3.90\\
MOCA~\cite{singh2020moca}  &22.05 &28.29& 15.10& 22.05 && 9.99 &14.28 &2.72 &5.30 \\
E.T.~\cite{pashevich2021episodic} &\textbf{34.93} & 45.44& 27.78& 38.42 && 11.46& 18.56& 4.10 &8.57\\
LWIT~\cite{nguyen2021look}&23.10 &40.53 &\textbf{43.10}& 30.92 && \textbf{16.34} &20.91& 5.60 &9.42\\
FILM~\cite{min2021film} &15.06& 38.51& 11.23& 27.67&& 14.30& 36.37 &\textbf{10.55} &26.49 \\
OPEx  &  22.08   & \textbf{54.81}   & 14.52  &  \textbf{44.03} & & 15.27  &  54.18 & 13.48 & \textbf{41.85}  \\
\hline
\multicolumn{9}{l}{\textbf{High-level goal instructions only}} \\
\hline
LAV~\cite{nottingham2021modular} & 13.18  &23.21 & 6.31  &13.35  & &10.47  &17.27  &3.12  &6.38\\
HLSM~\cite{blukis2022persistent}& 11.53 & 35.79 & 6.69 & 25.11  & & 8.45  &27.24  & 4.34  &16.29 \\
LGS-RPA~\cite{murray2022following} & \textbf{24.49}  &  41.71  & \textbf{16.65} &  33.01 & &  \textbf{20.01} & 38.55 & \textbf{12.92} & 27.80 \\
EPA~\cite{liu2022planning}&  3.47  &  44.14 & 2.56 &  39.96 & & 3.91  & 39.54 & 2.92 & 36.07 \\
LLM-Planner~\cite{song2023llm} & - & 24.57 &  - & 15.33 &  & - & 22.89 & - &  13.41 \\
FILM~\cite{min2021film}  & 14.17     & 36.15     & 10.39     & 25.77 & &  13.13  & 34.75 & 9.67  & 24.46\\
OPEx-S  &  20.13    &  \textbf{54.27}   &   13.64   & \textbf{43.51} & & 18.46  & \textbf{53.82} & 12.57 & \textbf{41.27} \\
\bottomrule[1.5pt]
\end{tabular}
\caption{Main Results on the test splits of ALFRED benchmark.}
\label{tab: OPEx_full_results}
\end{table*}

\section{Prompt Examples}
In this section, we provide three prompt examples for the LLM-based planner, LLM-based observer, and LLM-based executor respectively. 

\begin{figure}[htb]
    \centering
    \includegraphics[width=0.5\textwidth]{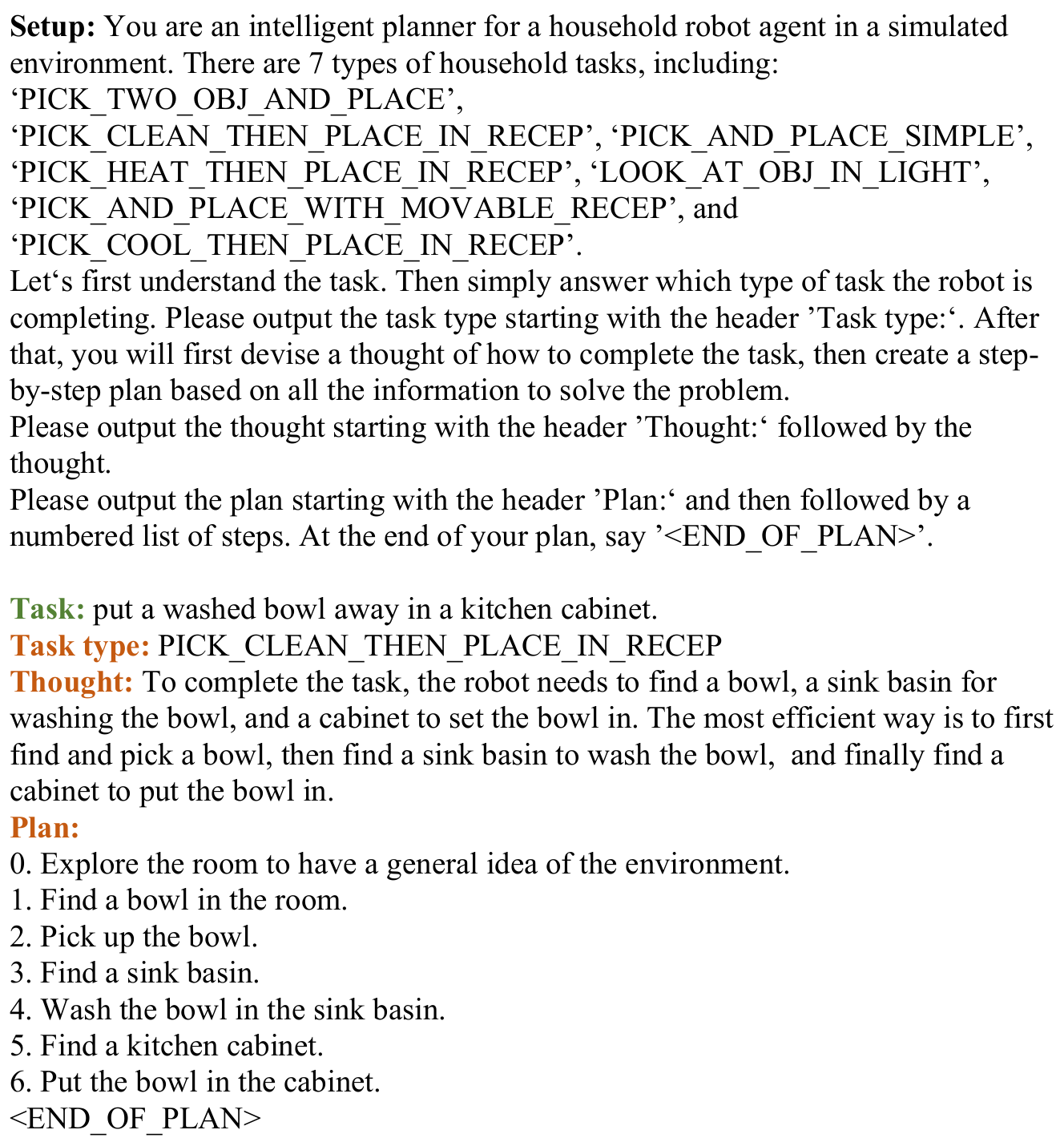}
    \caption{Prompt example of the LLM-based Planner. \textbf{Setup} is fixed for all the input test cases, \textcolor{ForestGreen}{\textbf{Task}} is the input to the LLM-based planner that varies for distinct input test cases, \textcolor{Bittersweet}{\textbf{Task type}}, \textcolor{Bittersweet}{\textbf{Tought}}, and \textcolor{Bittersweet}{\textbf{Plan}} are the content required to be generated by the LLM-based planner. The same color mode applies to other figures.}
    \label{fig:planner_prompt}
\end{figure}

\paragraph{LLM-based Planner.} 
In Figure \ref{fig:planner_prompt}, we present an illustrative prompt example of the LLM-based planner. The high-level instruction for this instance is "place a washed bowl into a kitchen cabinet." The prompt for the LLM-based planner is constructed to establish the planning task and define the desired output format. Specifically, the input provided to the planner is: "\textcolor{ForestGreen}{\textbf{Task:}} place a washed bowl into a kitchen cabinet." The resulting output encapsulates both the reasoning stages and the path of reasoning undertaken by the LLM-based planner. Given that the foundation of the planner's reasoning prowess lies in its comprehension, we initially expect it to demonstrate a fundamental understanding of the task. This is manifested through the presentation of the task's \textcolor{Bittersweet}{\textbf{Task type}} (in this instance, "PICK\_CLEAN\_THEN\_PLACE\_IN\_RECEP"). Subsequently, drawing inspiration from the concept of Chain-of-Thought Prompting, we introduce a two-step requirement. Firstly, the planner is prompted to generate its \textcolor{Bittersweet}{\textbf{Thought process}} in achieving the task, followed by the presentation of the ultimate \textcolor{Bittersweet}{\textbf{Plan}} to accomplish the specified task.

\begin{figure}[htb]
    \centering
    \includegraphics[width=0.5\textwidth]{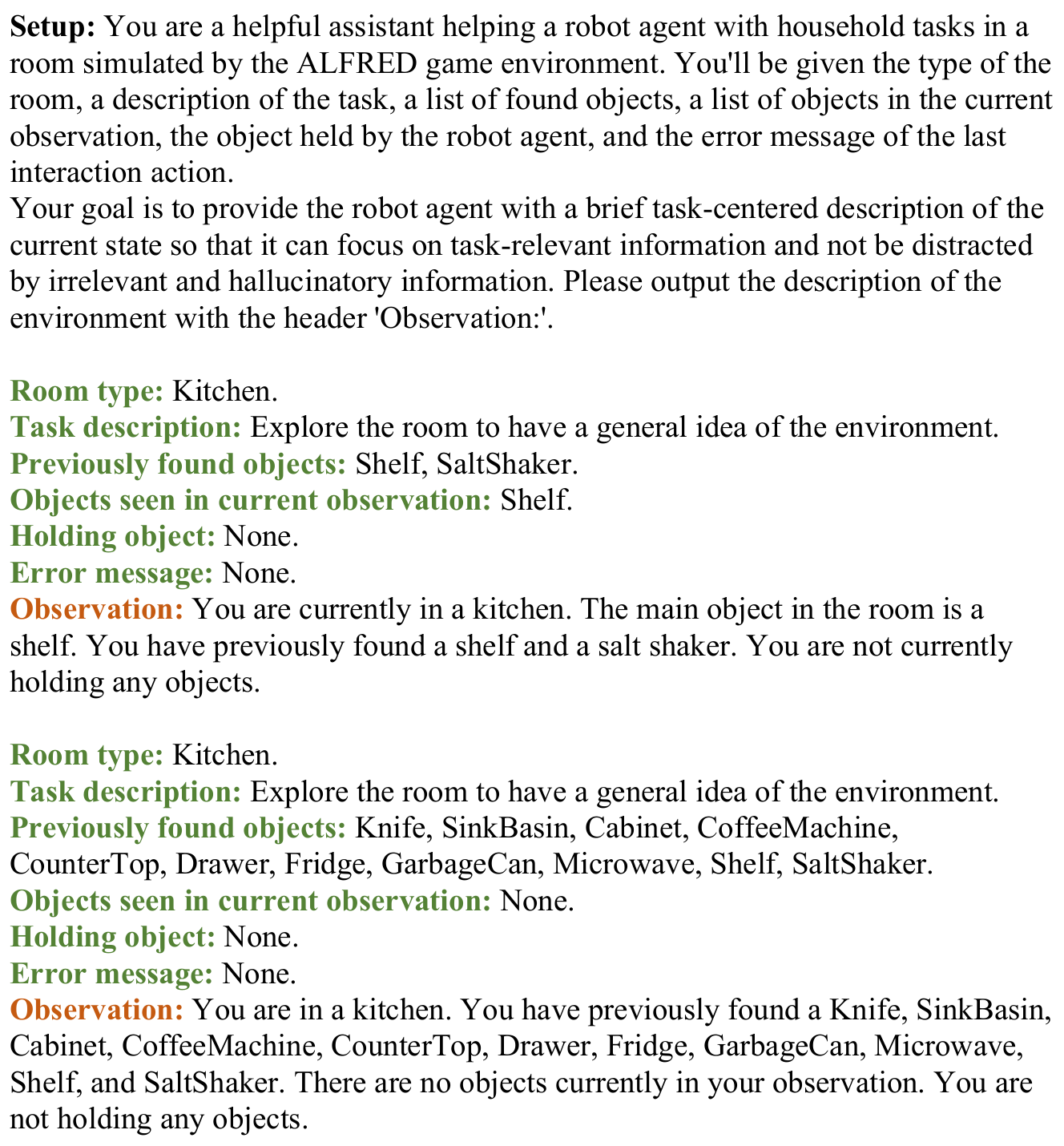}
    \caption{Prompt example of the LLM-based Observer.}
    \label{fig:observer_prompt}
\end{figure}

\paragraph{LLM-based Observer.} Fig. \ref{fig:observer_prompt} demonstrates two prompt examples for the LLM-based observer. Similar to the prompt design of the LLM-based planner, the prompt for the LLM-based observer also starts with a setup that establishes the observation task. The input to the observer is a set of information collected from the environment, including \textcolor{ForestGreen}{\textbf{Room type:}} indicating which kind of the room the agent is currently in (kitchen, living room, bedroom, or bathroom), \textcolor{ForestGreen}{\textbf{Task description:}} specifying the current subtask (which is generated by the LLM-based planner) to complete,  \textcolor{ForestGreen}{\textbf{Previously found objects:}} storing all the objects detected by the agent from the start of the episode to current time step, \textcolor{ForestGreen}{\textbf{Objects seen in current observation:}} pointing out the objects detected in the agent's current egocentric view,  \textcolor{ForestGreen}{\textbf{Holding object:}} tracking the object that is currently holden by the agent, and \textcolor{ForestGreen}{\textbf{Error message:}} tracking the error that causes action failures to facilitate exception handling capability of agent. Since successful action in the simulator typically results in the RGB change of the egocentric observation, we can detect action failures by comparing the egocentric observations before and after the execution of the action. If one kind of action failure is detected, then the error message of the corresponding action failure will be gathered by the LLM-centric observer. The designing purpose of the LLM-based observer is not only to gather information but also to serve as a ``information gate'' which filters out task-irrelevant information and effectively organizes the task-relevant information for better grounded planning and acting.

\begin{figure}[htb]
    \centering
    \includegraphics[width=0.5\textwidth]{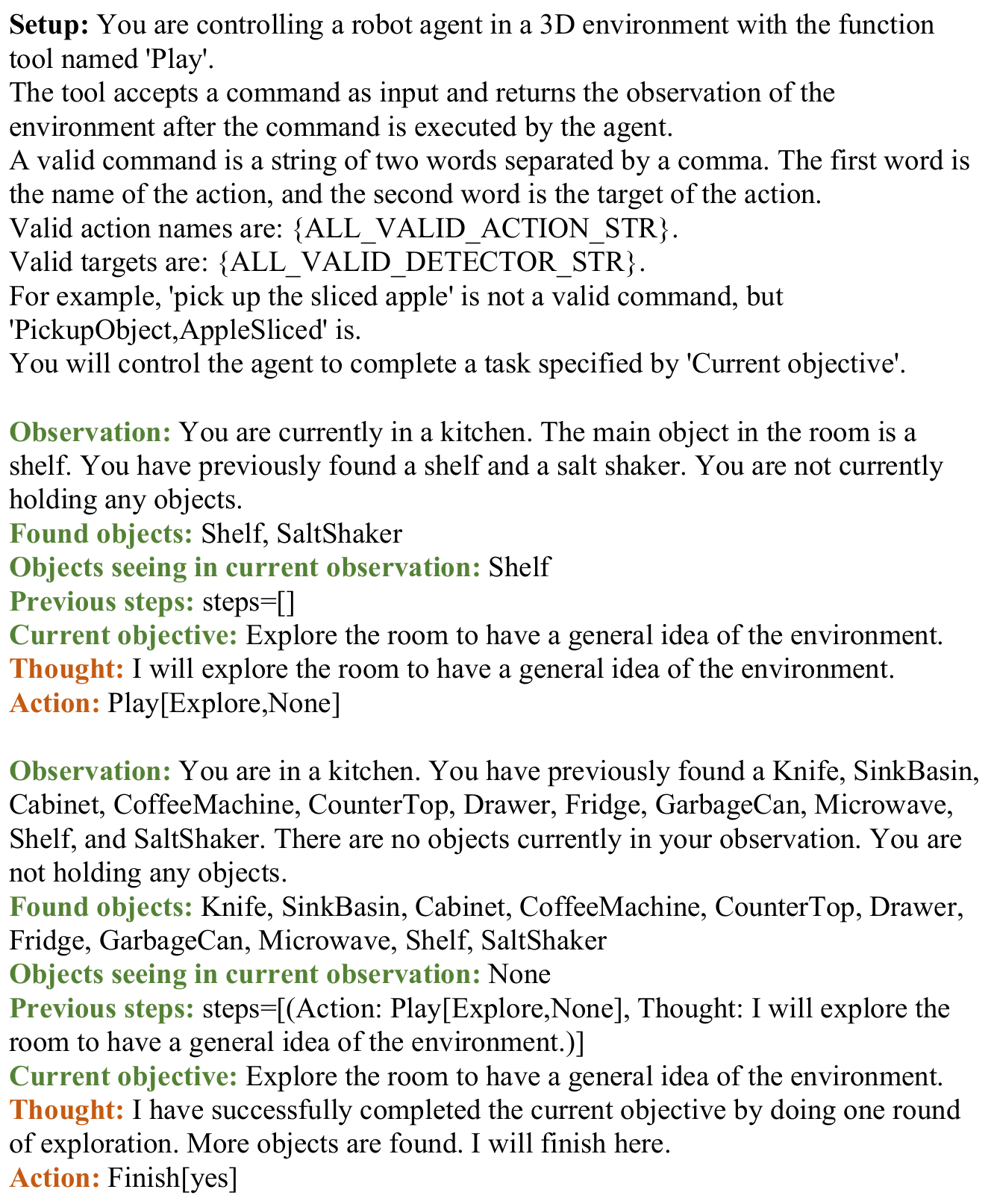}
    \caption{Prompt example of the LLM-based Executor.}
    \label{fig:executor_prompt}
\end{figure}

\paragraph{LLM-based Executor.} A prompt example of completing ``Explore the room to have a general idea of the environment'' is illustrated in Fig. \ref{fig:executor_prompt}. Specifically, the prompt of the LLM-based executor also starts with a setup establishing the execution task and indicating the desired output format. Afterward, the setup is followed by the input to the LLM-centric executor, which consists of \textcolor{ForestGreen}{\textbf{Observation:}} presenting the current language description of the word state generated by the LLM-based observer, \textcolor{ForestGreen}{\textbf{
Found objects:}} tracking all the objects detected by the agents, \textcolor{ForestGreen}{\textbf{
Objects seeing in current observation:}} noting the objects detected from current egocentric visual observation, \textcolor{ForestGreen}{\textbf{Previous steps:}} tracking the steps taken for the current subtask, and {\textbf{Current objective:}} specifying the current subtask to complete. Inspired by ReAct, we require the LLM-based executor to generate not only the final skill action plan \textcolor{Bittersweet}{\textbf{Action}} but also the reasoning paths \textcolor{Bittersweet}{\textbf{Thought}} in the first place.

\end{document}